\documentclass[11pt]{article}

\usepackage[preprint]{acl}

\usepackage{times}
\usepackage{latexsym}
\usepackage[T1]{fontenc}
\usepackage[utf8]{inputenc}
\usepackage{microtype}
\usepackage{inconsolata}
\usepackage{graphicx}
\usepackage{enumitem}
\usepackage{booktabs}
\usepackage{float} %

\usepackage{amsmath,amssymb}

\newif\ifarxiv
\newcommand{\arxivonly}[1]{\ifarxiv#1\fi}

\makeatletter
\def\endthebibliography{\def\@noitemerr{\@latex@warning{Empty `thebibliography' environment}}\endlist}
\makeatother

\title{HERD: Harness-Evolving Recursive Distillation for Continual Learning}

\arxivtrue

\newcommand{\aff}[1]{\textsuperscript{\normalfont#1}}
\newcommand{\authorline}{Xun Sun\aff{1} \quad Thanh Thi Nguyen\aff{2} \quad Thuy-Trang Vu\aff{1}}
\newcommand{\affnote}{\aff{1}Monash University \aff{2}University of the Sunshine Coast.
  Correspondence to: Xun Sun <xsun0087@student.monash.edu>,
  Thanh Thi Nguyen <tnguyen5@usc.edu.au>, Thuy-Trang Vu <trang.vu1@monash.edu>.}

\makeatletter
\AtBeginDocument{%
  \def\@maketitle{\vbox to \titlebox{\hsize\textwidth \linewidth\hsize \centering
    \vskip 0.1in
    \hrule height 1pt \vskip 0.22in
    {\Large\bfseries \@title \par}
    \vskip 0.22in \hrule height 1pt
    \vskip 0.25in
    {\bfseries \authorline \par}
    \vfil}}%
  \let\acl@maketitle\maketitle
  \def\maketitle{\acl@maketitle
    {\renewcommand\thefootnote{}\footnotetext{\raggedright\hspace*{-1.8em}\affnote}}%
    \setcounter{footnote}{0}}}
\makeatother

\begin{document}
\maketitle

\begin{abstract}
Continual learning requires a model to keep acquiring new skills and knowledge without compromising its existing capabilities.
Parametric methods suffer from catastrophic forgetting and loss of plasticity.
Non-parametric methods, such as prompts, skills, and harnesses, leverage in-context learning
but are limited by context length and lossy plain-text compression.
To effectively combine their complementary strengths, we propose HERD (Harness-Evolving Recursive Distillation), a hybrid continual learning framework that learns efficiently in the harness and internalizes experience into cumulative capabilities via distillation.
On the harness side, HERD continually refines its skills to capture reusable patterns, while retaining high-residual examples that the skills fail to capture in a bounded memory.
On the parameter side,
HERD performs drift-constrained distillation with dual anchors, substantially reducing forgetting over long recursive runs.
On the 25-stage task stream of TRACE+17, under fixed memory and compute budgets, HERD matches the strongest parametric baseline in new-task acquisition while exhibiting the least forgetting and the strongest retention of general capabilities.

\end{abstract}

\section{Introduction}
\label{sec:introduction}

\begin{figure*}[t]
    \centering
    \includegraphics[width=0.90\textwidth]{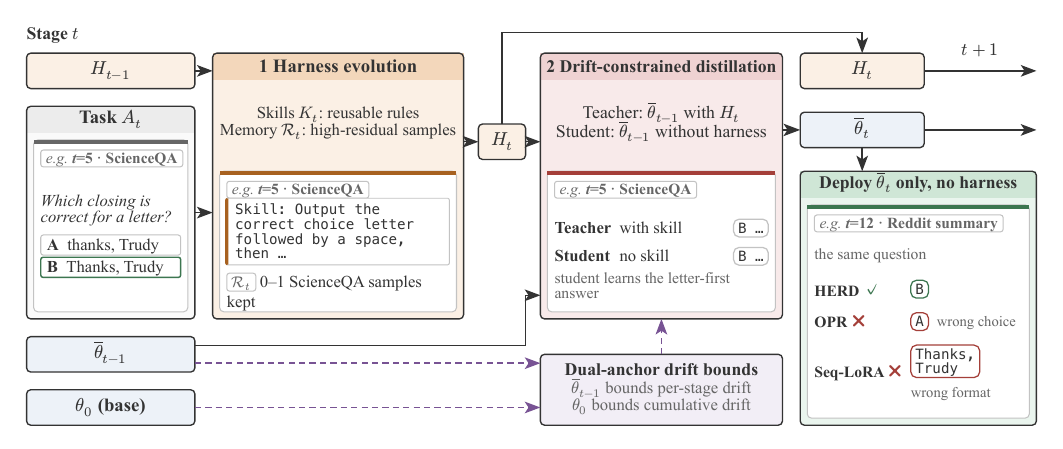}
    \vspace{-1em}
    \caption{HERD overview: one recursive stage. The harness \(H_t=(K_t,\mathcal R_t)\) evolves on task \(A_t\) and is distilled into the weights; drift budgets keep the student close to two anchors, the previous checkpoint \(\bar\theta_{t-1}\) and the base \(\theta_0\). \(H_t\) and \(\bar\theta_t\) recur, and only \(\bar\theta_t\) is deployed. The \emph{e.g.} tiles follow one ScienceQA item, learned at \(t=5\) and asked again at \(t=12\): two parametric baselines fail, OPR choosing the wrong option and Seq-LoRA writing the right text without the choice letter, while HERD answers from its weights alone across three seeds.}
    \label{fig:herd-overview}
\end{figure*}

Large language models (LLMs) are highly capable, yet cannot independently carry out much of the work people need done \citep{mazeika-etal-2025-remote-labor}.
One pressing problem in closing this gap is continual learning: humans keep learning from experience while retaining what they know, whereas models no longer change once deployed.
Continual learning (CL) seeks to enable this ability by learning from a stream of tasks without forgetting existing capabilities or retraining from scratch on earlier data~\citep{wu-etal-2024-continual}.

Parametric methods, however, inevitably encounter catastrophic forgetting \citep{mccloskey-cohen-1989-catastrophic,kirkpatrick-etal-2017-ewc,huang-etal-2024-mitigating} and loss of plasticity \citep{dohare-etal-2024-plasticity,yang-etal-2026-slora}.
They also give little control over what is internalized: raw experience carries noise, duplicates, and conflicts alongside the few samples that matter.

In-context learning (ICL) offers a non-parametric path to continual learning \citep{brown-etal-2020-language,gutierrez-etal-2025-rag-to-memory}.
New knowledge is stored in external memory and retrieved into context, so a frozen LLM learns across tasks \citep{shi-etal-2025-continual}.
This appears to sidestep the flaws above: a frozen model forgets nothing, loses no plasticity, and keeps knowledge inspectable and editable.
It is also the norm in practice: users write prompts, skills, and harnesses rather than post-train, and recent works build continual learning systems on harnesses \citep{zhang-etal-2026-ace,kang-etal-2026-hcl}.
Could a harness alone, recursively compressing its context, solve continual learning once models are strong and context windows long enough?
We argue that it cannot.
External stores can hold unbounded data, but nothing accumulates in the model itself: whatever it acts on must pass through the context, so experience is compressed into bounded plain text, which loses information at a very high rate.
The more experience accumulates, the more is lost.

The resolution lies in the weights themselves.
A few hundred gigabytes of parameters recall history, law, medicine, and science \citep{AL2024-knowledge3,pmlr-v306-morris26a,roberts-etal-2020-much}, or even reproduce entire books nearly verbatim \citep{cooper-etal-2026-extracting}.
Language models are general-purpose compressors \citep{deletang-etal-2024-compression}, and compression is intelligence \citep{hutter-2005-universal}.
Continual learning therefore still relies on weight updates, but should not apply them to every example.
Each layer should store what it is best suited for: capacity in the weights, selection in the harness.

We therefore propose HERD (Harness-Evolving Recursive Distillation), a hybrid continual learning system in which the harness is the fast, selective learning layer and the weights are the slow, persistent layer.
The design mirrors the human brain: a harness resembles episodic memory formed during the day, flexible but limited in capacity, which sleep consolidates into long-term memory \citep{mcclelland-etal-1995-complementary}.
HERD likewise internalizes its harness into the weights after every stage, under fixed memory and compute budgets and at no added inference cost.
It does not eliminate forgetting, but preserves existing capabilities as far as possible while the model keeps learning.

HERD is designed around recursion: it must run stably stage after stage, which raises two challenges.
First, the context is bounded, so the harness must decide what to change and what to keep.
Second, recursive training drifts, and unchecked drift ends in collapse.
HERD answers each with one module (Figure~\ref{fig:herd-overview}).
For the first, the harness evolves with the task stream: the model iterates on its skills to abstract reusable rules, and retains in a bounded memory the high-residual samples the skills cannot cover.
For the second, HERD performs drift-constrained distillation with dual anchors.
The student is the previous checkpoint trained to match itself under the harness, which shifts the model less than SFT and thus forgets less \citep{shenfeld-etal-2026-sdft,shenfeld-etal-2025-razor}.
The previous checkpoint bounds per-stage drift and the base model bounds cumulative drift.
The harness is kept, inspectable and editable, for the next stage; only the weights are deployed.

Across a 25-stage stream HERD keeps more of what it learns than any other method: its final mean over all seen tasks is 0.706, above every parametric baseline and 27 points above the non-parametric ones, while its acquisition, 0.713, is statistically indistinguishable from the plasticity floor set by Seq-LoRA (Table~\ref{tab:main-results}).
Forgetting is 0.010, about half or less that of every weight-updating baseline but SDFT, whose generation collapses (Appendix~\ref{sec:appendix-cases}); the general test falls 6.3 points where these other baselines lose 20.0 to 25.5.
In Figure~\ref{fig:herd-overview}, the rule learned at stage~5 still holds in the weights alone at stage~12 across three seeds.

Our contributions are: (1) HERD, a hybrid continual learning system that evolves a harness of skills and high-residual memory and recursively internalizes it into the weights (\S\ref{sec:methodology});
(2) drift-constrained distillation with dual anchors, where the previous checkpoint bounds per-stage drift and the base model bounds cumulative drift (\S\ref{sec:drift-constrained-opd});
(3) an evaluation on TRACE+17, a 25-stage stream of eight TRACE tasks \citep{wang-etal-2023-trace} and 17 Super-NaturalInstructions tasks \citep{wang-etal-2022-super}, against parametric, non-parametric, and hybrid baselines (\S\ref{sec:experiments}).

\arxivonly{Our code and dataset are available at \url{https://github.com/a1exsun/HERD}, and the project page is at \url{https://a1exsun.github.io/HERD}.}

\section{Methodology}
\label{sec:methodology}

\setcounter{dbltopnumber}{1}

\subsection{Preliminaries}
\label{sec:task-definition}
\label{sec:preliminaries}
\label{sec:problem-formulation}
\label{sec:harness-framework-formulation}

\paragraph{Paradigms.}
We consider an ordered stream of tasks $\mathcal{S}=(\mathcal{T}_1,\ldots,\mathcal{T}_T)$, each with its own splits and a metric $m_t$ in $[0,1]$.
Every example is an input and a target text; any dataset-specific instruction is part of the example.
The learner after stage $t$ is $\mathcal{L}_t=(f_t,M_t)$: a prediction model and everything it retains across stages, external indexes included.

\paragraph{Problem formulation.}
Stage $t$ exposes $E_t=(D_t^{\mathrm{train}},D_t^{\mathrm{val}})$ once and updates $\mathcal{L}_t=A(\mathcal{L}_{t-1},E_t)$ from $\mathcal{L}_0=(f_{\theta_0},M_0)$; earlier training sets are reachable only through $M$.
Retained state and update compute are bounded, $\mathrm{size}(M_t)\le B_M$ and $\sum_{t}\mathrm{cost}(A,\mathcal{L}_{t-1},E_t)\le B_C$, and any context attached at inference is reported with the scores.
The learner is scored on a fixed schedule on the tasks seen so far and on fixed general-capability probes.
With $R_{t,j}$ the score on task $j$ after stage $t$ and $\Delta C_k=C_{T,k}-C_{0,k}$ the change in the score on probe $k$ over the stream, we report final average performance, forgetting, and the worst change in general capability,
\begin{equation}
\begin{gathered}
    \mathrm{AA}=\tfrac{1}{T}\textstyle\sum_{j}R_{T,j},\quad
    \Delta C_{\min}=\min_{k}\Delta C_k,\\
    \mathrm{F}=\tfrac{1}{T-1}\textstyle\sum_{j<T}\max(0,\,R_{j,j}-R_{T,j}),
\end{gathered}
\label{eq:metrics}
\end{equation}
together with the mean change $\Delta C$ (\S\ref{sec:experiment-metrics}).
The objective is to maximize $\mathrm{AA}$ and minimize $\mathrm{F}$ without negative $\Delta C$ or $\Delta C_{\min}$, within $B_M$ and $B_C$.

\paragraph{Harness representation.}
We use harness in the sense of DeepSeek Harness (DSH) \citep{deepseek-ai-2026-harness} and of harness-level continual learning \citep{kang-etal-2026-hcl,zhang-etal-2026-self-harness}: the state and procedures around a model at inference, such as system prompts, skills, memories, and tools, as opposed to its weights.
The harness of HERD is $H=(K,\mathcal{R})$.
$K$ is a registry of skills, each a routing prefix and instructions; an input $x$ is routed by prefix, and the matched instructions $K(x)$ become the system message.
$\mathcal{R}$ is a residual memory of whole training records, prompt and verified answer, kept verbatim.
The budgets are $B_K$ and $B_R$, with $B_M=B_K+B_R$.
Executing the harness is $f(x;\theta,K)$: the model $\theta$ answering $x$ under the system message $K(x)$.

\subsection{HERD in One Stage}
\label{sec:hero-overview}

In this notation, HERD after stage $t$ is $\mathcal{L}_t=(f_{\bar\theta_t},(K_t,\mathcal{R}_t,G))$: merged dense weights $\bar\theta_t$, the harness $H_t=(K_t,\mathcal{R}_t)$, and a fixed set $G$ of general probes reserved before the stream.
Figure~\ref{fig:herd-overview} shows one stage, which learns the task twice.
First, the harness evolves on a bounded window of the task (\S\ref{sec:recursive-bounded-harness}).
Second, the checkpoint, given the skill and the verified answer, teaches a fresh adapter that sees only the bare prompt, under drift budgets toward $\bar\theta_{t-1}$ and $\theta_0$ (\S\ref{sec:drift-constrained-opd}); the adapter is merged into $\bar\theta_t$, the memory is refreshed under the merged weights, and the harness is carried on.
Both steps are driven by the residual of a record (Eq.~\ref{eq:residual}).

\begin{figure*}[t]
    \centering
    \includegraphics[width=\textwidth]{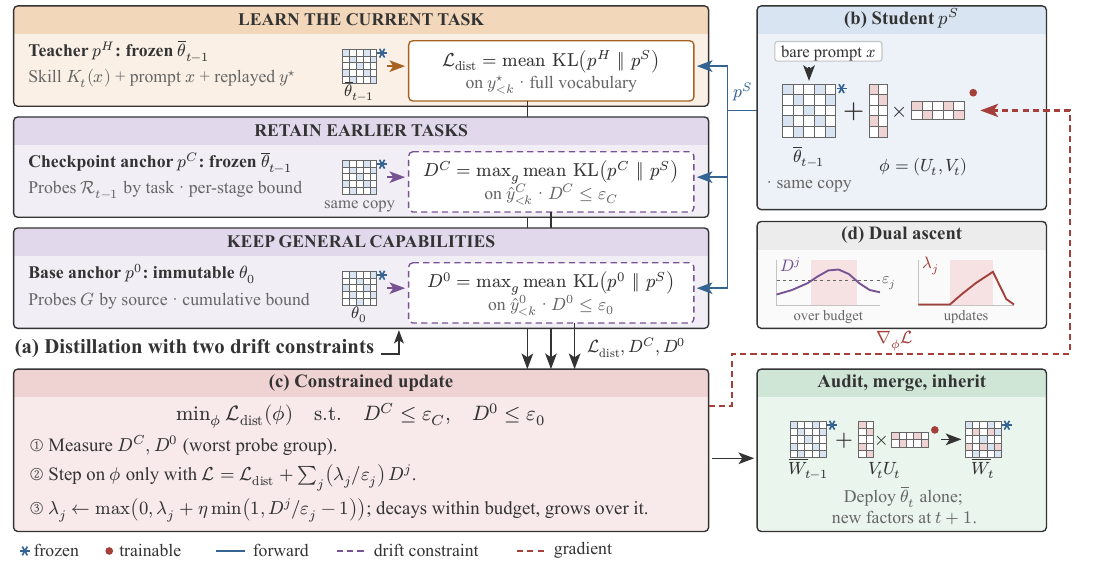}
    \caption{Drift-constrained distillation. (a) One frozen copy of \(\bar\theta_{t-1}\) serves as both the teacher \(p^H\) and the checkpoint anchor \(p^C\); the immutable \(\theta_0\) is the general anchor \(p^0\). (b) The student adds fresh factors \(\phi\) to the same weights and sees only the bare prompt. (c) Each update measures both drifts, steps on \(\phi\), and moves the multipliers by dual ascent. (d) A multiplier grows only while its budget is exceeded.}
    \label{fig:drift-constrained-opd}
\end{figure*}

\subsection{Harness Evolution}
\label{sec:recursive-bounded-harness}

\paragraph{Proposal.}
The harness reads the task through a bounded window $A_t\subset D_t^{\mathrm{train}}$ of whole records filled to a budget $B_A$, the bounded context of \S\ref{sec:introduction}: it must decide what to change and what to keep.
The frozen checkpoint first answers $A_t$ without a skill; its wrong answers go to the skill writer, so the skill targets what the model now gets wrong rather than restating the task.
The base $\theta_0$ then writes the skill in one greedy call from the window and the mistakes.
Keeping the writer at $\theta_0$ keeps it fixed across the recursion; co-evolution with the weights is left to future work (Appendix~\ref{sec:appendix-future-work}).

\paragraph{Guard.}
The residual $r(x,y^*;\theta,K)$ of a record under a model $\theta$ executing a registry $K$ is the mean negative log-likelihood of its verified answer with the skill, but not the answer, in context,
\begin{equation}
    r = -\frac{1}{|y^*|}\sum_{k=1}^{|y^*|}\log p(y^*_k \mid K(x),x,y^*_{<k};\theta).
    \label{eq:residual}
\end{equation}
A candidate registry is committed only if the mean residual of $A_t$ under $\bar\theta_{t-1}$ is lower than under $K_{t-1}$; otherwise $K_t=K_{t-1}$: a skill survives only if it helps the checkpoint predict the task.

\paragraph{Residual refresh.}
After distillation and merge, the residual of every record in $\mathcal{R}_{t-1}\cup A_t$ is recomputed under $\bar\theta_t$ executing $K_t$, and $\mathcal{R}_t$ keeps whole records in decreasing residual until $B_R$ is filled.
What the skill and the new weights together still cannot predict stays in plain text; with one global budget, a record the weights have just absorbed yields its place at once.
$\mathcal{R}$ is never replayed and never a training target.
It supplies the probes of the next stage's checkpoint-drift constraint: on the records the new checkpoint finds hardest, the student may not leave it.

\subsection{Drift-Constrained Distillation}
\label{sec:drift-constrained-opd}

\paragraph{Teacher and student.}
The teacher $p^H$ is the frozen $\bar\theta_{t-1}$ under a privileged context: the routed skill $K_t(x)$ as the system message and the verified answer $y^*$ replayed as the model's own previous turn.
The student $p^S$ is the same weights with a freshly initialized LoRA adapter \citep{hu-etal-2022-lora} and sees only the bare prompt (Figure~\ref{fig:drift-constrained-opd}a,b).
The context carries no wording about the answer, so the student learns the answer and not the scaffolding around it.

With $p^H_k=p^H(\cdot\mid K_t(x),x,y^*,y^*_{<k})$ and $p^S_k=p^S(\cdot\mid x,y^*_{<k})$, the student minimizes the full-vocabulary forward KL divergence at every position of the verified answer,
\begin{equation}
    \mathcal{L}_{\mathrm{dist}} = \mathbb{E}_{(x,y^*)}\,\frac{1}{|y^*|}\sum_{k=1}^{|y^*|}
    \mathrm{KL}\!\left(p^H_k \,\|\, p^S_k\right).
    \label{eq:distill}
\end{equation}
Scoring the verified answer's own prefix rather than student samples closes degenerate paths such as echoing the prompt, on which teacher and student agree and the gradient vanishes.

\paragraph{Drift.}
A probe is a prompt $x$ with the first $n$ greedy tokens $\hat y$ of an anchor $a$; with $p^a_k=p^a(\cdot\mid x,\hat y_{<k})$ and $p^S_k$ likewise, drift over probe groups $P_g$ is the worst group's mean divergence,
\begin{equation}
    D(P;a) = \max_{g}\,\mathbb{E}_{P_g}\,\frac{1}{n}\sum_{k=1}^{n}
    \mathrm{KL}(p^{a}_k \,\|\, p^S_k).
    \label{eq:drift}
\end{equation}
Only the opening is scored because it decides the mode of a response: the failures in Figure~\ref{fig:herd-overview}, a wrong option and a dropped choice letter, are flips of the first few tokens that a long average would hide.
The worst group is taken because a collapse confined to one task, Seq-LoRA's correct ScienceQA answers falling from 94 to 15 across the stages of the same figure, barely moves a mean over all probes, which is also why we report $\Delta C_{\min}$.

\paragraph{Two anchors.}
Checkpoint drift $D^C$ anchors $\bar\theta_{t-1}$ on probes from $\mathcal{R}_{t-1}$, grouped by task, and bounds per-stage drift and the recently learned tasks; base drift $D^0$ anchors $\theta_0$ on $G$, grouped by source, and bounds cumulative drift and general capability.
Each bounds what the other cannot: the previous checkpoint keeps every step small, but small steps accumulate; the base cannot protect a task it never knew.
The budgets $\varepsilon_C$ and $\varepsilon_0$ are in units of divergence, so one pair serves every stage, which a penalty coefficient could not since the scale of the loss changes from stage to stage.

\paragraph{Enforcement.}
The stage solves
\begin{equation}
    \min_{\phi}\;\mathcal{L}_{\mathrm{dist}}(\phi)\;\;\text{s.t.}\;\; D^C\le\varepsilon_C,\; D^0\le\varepsilon_0,
    \label{eq:constrained}
\end{equation}
over the adapter parameters $\phi$ with multipliers $\lambda_C,\lambda_0\ge 0$: each update measures both drifts, adds every active constraint's worst-group divergence to the loss with weight $\lambda_j/\varepsilon_j$, steps on $\phi$ only, and moves the multipliers by bounded dual ascent,
\begin{equation}
    \lambda_j \leftarrow \max(0,\,\lambda_j + \eta\min(1,\,D^j/\varepsilon_j - 1)),
    \label{eq:dual}
\end{equation}
so that within budget a multiplier decays to zero and leaves learning alone, and over budget it grows and pulls the student back.
The final adapter's drift is audited against the original budgets (Figure~\ref{fig:drift-constrained-opd}c).

\paragraph{One checkpoint, three roles.}
One copy of the previous checkpoint is resident as the student's backbone, the teacher $p^H$, its forward pass with the adapter disabled, and the anchor $p^C$; both anchors' states on their fixed probes are cached.
With a fixed number of updates and probes per stage, the cost does not grow with $t$; only $\bar\theta_t$ is deployed, and budgets and constants are listed in Appendix~\ref{sec:appendix-implementation}.

\section{Experiments}
\label{sec:experiments}

\setcounter{dbltopnumber}{1}

\begin{table*}[t]
    \centering
    {\small%
\setlength{\tabcolsep}{4pt}
\begin{tabular}{llccccccc}
\toprule
Method & Family & AA$\uparrow$ & Acq.$\uparrow$ & F$\downarrow$ & BWT$\uparrow$ & G-test$\uparrow$ & $\Delta C\uparrow$ & $\Delta C_{\min}\uparrow$ \\
\midrule
\textbf{HERD} & This work & \textbf{0.706} & 0.713 & 0.010 & $-$0.007 & 0.774 & $-$0.063 & $-$0.130 \\
\midrule
Base & Control & 0.419 & 0.419 & 0.000 & 0.000 & 0.837 & 0.000 & 0.000 \\
Static harness & Control & 0.432 & 0.432 & 0.000 & 0.000 & 0.791 & $-$0.046 & $-$0.135 \\
\midrule
HCL-Stable & Context & 0.433 & 0.435 & 0.003 & $-$0.002 & \textbf{0.823} & \textbf{$-$0.014} & \textbf{$-$0.085} \\
ACE & Context & 0.427 & 0.428 & \textbf{0.001} & \textbf{$-$0.001} & 0.748 & $-$0.089 & $-$0.320 \\
\midrule
Seq-LoRA & Weights & 0.694 & \textbf{0.715} & 0.025 & $-$0.022 & 0.597 & $-$0.240 & $-$0.665 \\
Uniform replay & Weights & 0.699 & 0.711 & 0.020 & $-$0.013 & 0.599 & $-$0.238 & $-$0.675 \\
OPR & Weights & 0.690 & 0.705 & 0.024 & $-$0.016 & 0.614 & $-$0.223 & $-$0.625 \\
TreeLoRA & Weights & 0.686 & 0.711 & 0.032 & $-$0.026 & 0.605 & $-$0.232 & $-$0.635 \\
MoRAM & Weights & 0.693 & 0.701 & 0.019 & $-$0.009 & 0.637 & $-$0.200 & $-$0.650 \\
SDFT & Weights & 0.021 & 0.024 & 0.003 & $-$0.003 & 0.172 & $-$0.665 & $-$0.770 \\
\midrule
OPHSD & Hybrid & 0.620 & 0.639 & 0.021 & $-$0.020 & 0.582 & $-$0.255 & $-$0.645 \\
\bottomrule
\end{tabular}
}
    \caption{TRACE+17 results after stage 25 under a matched model, stream, splits, prompts, and schedule; means over three seeds. Bold: best learning method (controls excluded) per column. BWT: mean$_{j<T}(R_{T,j}-R_{j,j})$.}
    \label{tab:main-results}
\end{table*}

Our experiments evaluate HERD under a unified protocol to answer three questions.
\textbf{Q1.} Does HERD outperform purely parametric and purely non-parametric methods?
\textbf{Q2.} As the number of CL stages grows, does HERD maintain long-horizon stability?
\textbf{Q3.} Are harness evolution, high-residual memory, and the two drift anchors all indispensable?

\begin{figure*}[t]
    \centering
    \includegraphics[width=\textwidth]{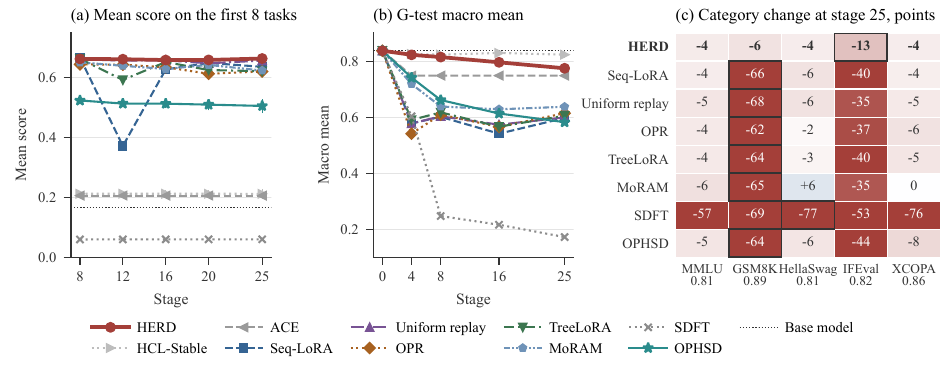}
    \vspace{-1em}
    \caption{Long-horizon stability on TRACE+17. (a) Mean score on tasks 1 to 8 at the retention stages. (b) G-test macro mean at the anchor stages; error bars are one standard deviation over three seeds, the dotted line is the base model. (c) Change in each G-test category after stage 25 against the base model, in points, for the weight-updating methods; the outlined cell is each method's $\Delta C_{\min}$. Per-stage trajectories and HERD's drift audit are in Figure~\ref{fig:long-horizon-appendix}.}
    \label{fig:experimental-summary}
\end{figure*}

\subsection{Experimental Settings}
\label{sec:unified-protocol}

\paragraph{Benchmark.}
TRACE+17 appends 17 Super-NaturalInstructions tasks \citep{wang-etal-2022-super} to the eight TRACE tasks \citep{wang-etal-2023-trace}, because eight stages are too few to tell stability from slow collapse.
Each of the 25 tasks has 500 training, 100 validation, and 100 test examples; no prompt appears in more than one task or split.
A G-test of five categories, 200 held-out items each, tracks general capability: knowledge, mathematics, commonsense, instruction following, and multilingual understanding (MMLU, GSM8K, HellaSwag, IFEval, XCOPA; \citealp{hendrycks-etal-2020-mmlu,cobbe-etal-2021-gsm8k,zellers-etal-2019-hellaswag,zhou-etal-2023-ifeval,ponti-etal-2020-xcopa}).
All methods start from Qwen3.5-4B with the same decoding, length limits, and scorers; all weight updates share one budget, 32 updates of 32 records per stage.
After every stage we score the current task; after stages 4, 8, 16, and 25 we also score all seen tasks and the G-test (Appendix~\ref{sec:appendix-experiments}).

\paragraph{Baselines.}
\label{sec:comparison-methods}
\emph{Non-parametric}: ACE \citep{zhang-etal-2026-ace} grows a playbook; HCL \citep{kang-etal-2026-hcl} keeps a harness edit only if it improves the current task and breaks no retained example; both attach the harness to every prompt at inference.
\emph{Parametric}: Seq-LoRA trains one adapter through the stream and serves as the floor; uniform replay keeps a memory the size of HERD's residual memory, filled by uniform sampling; OPR \citep{chen-etal-2026-opr} replays the model's own best earlier answers; TreeLoRA \citep{qian-etal-2025-treelora} and MoRAM \citep{lu-etal-2026-moram} isolate updates, in a gradient-similarity tree and in rank-1 experts; SDFT \citep{shenfeld-etal-2026-sdft} self-distills from a demonstration-conditioned teacher, without a harness.
\emph{Hybrid}: OPHSD \citep{zhao-etal-2026-ophsd} distills a fixed harness into the weights and deploys without it; the harness never changes and drift is not budgeted.
\emph{Controls}: the base model as is, and the same model under a fixed harness that never learns.

\paragraph{Metrics.}
\label{sec:experiment-metrics}
Acquisition, the mean of $R_{j,j}$, is each task's score right after it is learned; AA is what remains after stage 25; F is what is lost in between (Eq.~\ref{eq:metrics}).
We report both the mean change over the five G-test categories, $\Delta C$, and the worst, $\Delta C_{\min}$, because a collapse in one category can hide behind a mean of five (\S\ref{sec:drift-constrained-opd}).
All numbers are means over three seeds, with standard deviations and 95\% bootstrap intervals over items.

\subsection{Main Results}
\label{sec:experiment-results}

Table~\ref{tab:main-results} reports every method after stage 25 and Figure~\ref{fig:experimental-summary} the trajectories behind it.
HERD lies on the frontier of the three families: it acquires as much as the strongest parametric method, retains more than any method that updates weights, loses the least general capability among them, and deploys without context.

\paragraph{Non-parametric methods.}
ACE and HCL finish within 1.4 points of the unchanged base model (AA 0.427 and 0.433 against 0.419): the stream leaves almost no trace in their harnesses.
ACE's playbook reaches 80 bullets and 15,410 tokens by task 2 and stops growing; once it fills the prompt, nearly every later training call exceeds the 16,384-token input ceiling and is rolled back, yet every test question still carries those 15,410 tokens of context.
HCL commits nearly none of its proposals in 25 stages, so its scores amount to the base model under a fixed system prompt.
HERD, scored bare, is 27 points above both; the paired 95\% intervals against both exclude zero (Table~\ref{tab:comparisons}).

\paragraph{Parametric methods.}
HERD's acquisition, 0.713, is statistically indistinguishable from Seq-LoRA's 0.715, the plasticity floor (paired difference $-0.002$, interval $[-0.010, 0.007]$), and from uniform replay and TreeLoRA; it is 0.8 points above OPR, 1.2 above MoRAM, and 7.4 above OPHSD.
Its AA of 0.706 is the highest in the table, significantly above Seq-LoRA, OPR, TreeLoRA, MoRAM, and OPHSD (1.2 to 8.7 points) and tied with uniform replay (0.7 points, interval $[-0.001, 0.016]$).
Forgetting is 0.010, under half of Seq-LoRA's, OPR's, TreeLoRA's, and OPHSD's and about half of uniform replay's and MoRAM's, with paired intervals excluding zero against Seq-LoRA, uniform replay, OPR, TreeLoRA, MoRAM, and OPHSD (except SDFT's 0.003 accompanies an AA of 0.021).
SDFT's F of 0.003 reflects collapse, not retention: it acquires only 0.024 (Appendix~\ref{sec:appendix-cases}).
Uniform replay holds exactly as many tokens as HERD's residual memory and still forgets like the others (F 0.020, $\Delta C$ $-0.238$): the gap is not in how much is stored, and \S\ref{sec:ablation-studies} locates it in what is kept and what is constrained.

\paragraph{General capability.}
The parametric baselines other than SDFT pay for new tasks with old abilities: $\Delta C$ is $-0.20$ to $-0.24$, and the worst category falls by 62 to 68 points.
The loss concentrates in GSM8K and IFEval, which collapse within four stages (Seq-LoRA: from 0.89 and 0.82 to 0.18 and 0.33) and never recover, while knowledge, commonsense, and multilingual scores stay within a few points of the base (Figure~\ref{fig:experimental-summary}c).
HERD's $\Delta C$ of $-0.063$ and $\Delta C_{\min}$ of $-0.130$ are the smallest of the weight-updating methods by a wide margin: every parametric baseline trails its G-test macro average by 13.7 to 60.2 points, and the five parametric baselines with paired bootstrap comparisons in Table~\ref{tab:comparisons} by 13.7 to 17.7 points; its largest loss, 13 points, falls in the same instruction-following category.

\paragraph{Cost.}
A HERD stage runs a fixed number of updates and probes, so its training time tracks answer length, not the stage index: 316~s at stage 1, 284~s at stage 25, 10,723~s in total against Seq-LoRA's 10,463~s.
OPHSD, which regenerates its teacher context by rollouts at every stage, needs 659,000~s; ACE pays its 15,410 tokens at every inference.

\paragraph{Long-horizon stability.}
On the first eight tasks (Figure~\ref{fig:experimental-summary}a), HERD's mean stays within half a point of its stage-8 level throughout, from 0.664 at stage 8 to 0.665 at stage 25.
Every parametric baseline moves more: Seq-LoRA falls to 0.372 at stage 12 before recovering to 0.637, TreeLoRA to 0.593, and OPR and MoRAM decline to 0.621 and 0.625.
On the G-test, the parametric baselines take most of their loss by stage 4 or 8 and then plateau, while HERD loses 1.5 points by stage 4 and 6.3 by stage 25.
The two recursive self-distillation baselines instead drift further at every checkpoint.
OPHSD's G-test falls through 0.741, 0.661, 0.613, and 0.582 at stages 4, 8, 16, and 25, and its first eight tasks from 0.524 to 0.506.
SDFT, whose teacher is asked to include its thinking process, produces answers that lengthen stage after stage, from 345 tokens on average at stage 1 to the 5,120-token ceiling from stage 17 on; 14,098 of its 17,200 test predictions are cut off, and it ends at AA 0.021 and G-test 0.172.
HERD also distills the model from itself under a privileged context, but its teacher context is a skill and the verified answer rather than text generated by the drifting model, its loss is scored on the verified answer, and two anchors bound drift (\S\ref{sec:drift-constrained-opd}): the audited stage-25 drifts are 0.021 and 0.041 against budgets of 0.05.

\subsection{Ablation Study}
\label{sec:ablation-studies}

\begingroup
\setlength{\intextsep}{2pt}
\setlength{\abovecaptionskip}{4pt}
\begin{figure}[H]
    \centering
    \includegraphics[width=\columnwidth,trim=0 16bp 0 10bp,clip]{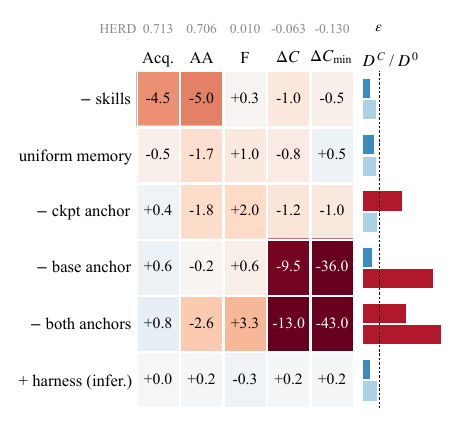}
    \caption{Ablations after stage 25: changes from full HERD in points (red is worse; colors clipped at $\pm10$). The strip shows audited drifts $D^C$ (upper) and $D^0$ (lower) against $\varepsilon=0.05$. Single seed; full table in Appendix~\ref{sec:appendix-ablation}.}
    \label{fig:hero-ablation}
\end{figure}
\endgroup

Figure~\ref{fig:hero-ablation} removes or replaces one component of HERD at a time and reports the change after stage 25; Table~\ref{tab:hero-ablation} adds the SFT objective and the $\varepsilon_C$ sweep.
Each component protects one quantity.

\paragraph{Harness evolution.}
Without skills, the teacher sees only the verified answer and nothing is proposed or committed; memory, probes, and anchors are unchanged.
Acquisition falls by 4.5 points and AA by 5.0, the largest losses in both columns, while forgetting and general capability move by at most a point.
Without both anchors, the harness evolves and nothing constrains the weights: acquisition holds, and retention and general capability are those of a parametric baseline.
OPHSD, whose harness never changes and whose drift is never budgeted, has lower acquisition and a larger general-capability loss than HERD (Acq.\ 0.639, $\Delta C_{\min}$ $-0.645$).
Only the full model, which evolves the harness under constraint, is on the frontier.

\paragraph{High-residual memory.}
Filling the memory by uniform sampling under the same budget leaves acquisition within half a point, but forgetting doubles, from 0.010 to 0.020, the value of uniform replay in Table~\ref{tab:main-results}.
The checkpoint probes are drawn from the memory, so uniform selection anchors the checkpoint on records the weights already predict and leaves the hardest earlier records unguarded; the stage-end $D^C$ stays within budget at 0.034.

\paragraph{Two anchors.}
Without the checkpoint anchor, forgetting rises to 0.030, close to the worst parametric baseline's 0.032, and the audited $D^C$ ends at 0.118, more than twice the budget.
Without the base anchor, the checkpoint constraint remains active, but base drift accumulates: the stage-end $D^0$ grows from 0.104 at stage 4 to 0.212 at stage 25, $\Delta C_{\min}$ falls from $-0.031$ to $-0.490$, and mathematics drops from 0.885 to 0.395 where the full model keeps 0.823.
With neither anchor, forgetting reaches 0.044, above every parametric baseline, and $\Delta C_{\min}$ reaches $-0.560$, close to their $-0.63$ to $-0.68$.

\paragraph{Objective and budget.}
Replacing distillation by cross-entropy on the verified answers, under the same adapter, budget, and constraints, doubles the general loss, to $\Delta C$ $-0.123$ and $\Delta C_{\min}$ $-0.320$, with both audited drifts within budget; this suggests that the teacher distribution limits drift where the probes do not reach.
The budget $\varepsilon_C$ trades forgetting against acquisition in the expected direction (Appendix~\ref{sec:appendix-ablation}).
Attaching the stage-25 harness at inference reduces forgetting by 0.3 points and changes the other metrics by at most 0.2 points (last row).

\section{Related Work}
\label{sec:related-work}

\paragraph{Parametric continual learning.}
Surveys group LLM continual learning by training stage \citep{wu-etal-2024-continual,shi-etal-2025-continual}; we group it by where experience is kept.
Structural methods isolate task updates in low-rank subspaces, adapter trees, per-task experts, or retained prompts and modules \citep{wang-etal-2023-olora,yang-etal-2026-slora,qian-etal-2025-treelora,lu-etal-2026-moram,razdaibiedina-etal-2023-progressive,zhao-etal-2024-sapt}.
Replay methods synthesize or select the data mixed into training \citep{huang-etal-2024-mitigating,chen-etal-2026-opr,feng-etal-2026-forever,lee-etal-2026-oasis}.
Self-supervised methods write their own fine-tuning data \citep{zweiger-etal-2025-seal}, distill from a demonstration-conditioned self \citep{shenfeld-etal-2026-sdft}, or consolidate fast memories into stable weights \citep{behrouz-etal-2026-sleep}.
Other work finds part of measured forgetting spurious \citep{zheng-etal-2025-spurious}, consolidates in weight space \citep{cho-etal-2026-forget-forgetting}, or reports limits of on-policy self-distillation \citep{wang-etal-2026-denser}.
HERD also updates low-rank factors by self-distillation, but a harness decides what to internalize and drift budgets bound each step.

\paragraph{Non-parametric continual learning.}
Non-parametric methods adapt an external memory, context, or harness around a fixed model, where the stability--plasticity tension reappears through retrieval \citep{gutierrez-etal-2025-rag-to-memory,hu-etal-2026-cl-memory}.
Recent systems consolidate memory banks \citep{ye-etal-2026-auto-dreamer}, evolve contexts into playbooks \citep{zhang-etal-2026-ace}, gate harness edits on improvement and retention checks \citep{kang-etal-2026-hcl}, or refine harnesses over task streams \citep{wei-etal-2026-evo-harness,liu-etal-2026-adaptive-harness}.
All pass experience through a bounded context, so nothing accumulates in the model itself (\S\ref{sec:introduction}).
HERD keeps such a harness as its fast layer but internalizes it into the weights after every stage.

\paragraph{Hybrid continual learning.}
A few methods combine the two.
OEL consolidates deployment experience online \citep{ye-etal-2026-oel}, Continual Harness updates the model from teacher-relabeled harness rollouts \citep{karten-etal-2026-continual-harness}, and EvoSC consolidates trajectories into soft prompts \citep{yu-etal-2026-self-consolidation}.
HERD keeps only skills and the high-residual samples they miss, and deploys the weights alone.

\paragraph{Distillation and harness adaptation.}
Context distillation internalizes instructions and examples into the weights \citep{snell-etal-2022-context}.
On-policy variants condition the teacher on experiential knowledge \citep{ye-etal-2026-opcd} or on a harness \citep{zhao-etal-2026-ophsd}, and RL's Razor ties forgetting to the distributional shift that on-policy training keeps small \citep{shenfeld-etal-2025-razor}.
HERD's privileged teacher is of this kind, but it keeps the harness across stages where OPHSD discards it, bounds drift explicitly, and scores verified answer prefixes instead of student rollouts.
Self-Harness accepts edits only after regression testing on fixed benchmarks \citep{zhang-etal-2026-self-harness}; HERD's commit gate is a simpler version of the same check (\S\ref{sec:recursive-bounded-harness}).
Two evaluations separate updating a harness from benefiting from it \citep{lin-etal-2026-harness-benefit} and ask whether optimizer gains compound over a continual stream \citep{wang-etal-2026-agent-optimizers}.

\section{Conclusion}
\label{sec:conclusion}

We presented HERD, a hybrid continual learning system that learns in the harness and internalizes it into the weights.
The harness evolves with the stream, iterating on its skills to abstract reusable patterns and keeping the high-residual samples they do not cover in a bounded memory.
Distillation then internalizes the harness under drift constraints with two anchors, the previous checkpoint for per-stage drift and the base model for cumulative drift; only the weights are deployed.
On the 25-stage TRACE+17 stream, HERD matches the strongest parametric baseline in acquisition, forgets the least among weight-updating methods that avoid collapse, and loses the least general capability.
Removing any part costs acquisition, retention, or general capability, and attaching the final harness at inference barely changes the scores.

\section*{Limitations}
Our empirical validation is currently scoped to one base model and one 25-stage stream rather than a large-scale sweep.
HERD also relies on a verified answer $y^*$ for every training record and assumes a stream with known task boundaries, one task per stage, and no revisits.
A detailed discussion regarding these boundaries and the corresponding plans is provided in Appendix~\ref{sec:appendix-future-work}.

\bibliography{custom}

\clearpage
\appendix
\makeatletter
\newcommand{\appendixfigcaption}{\def\@captype{figure}}
\makeatother
\twocolumn[{%
    \centering
    \includegraphics[width=\textwidth]{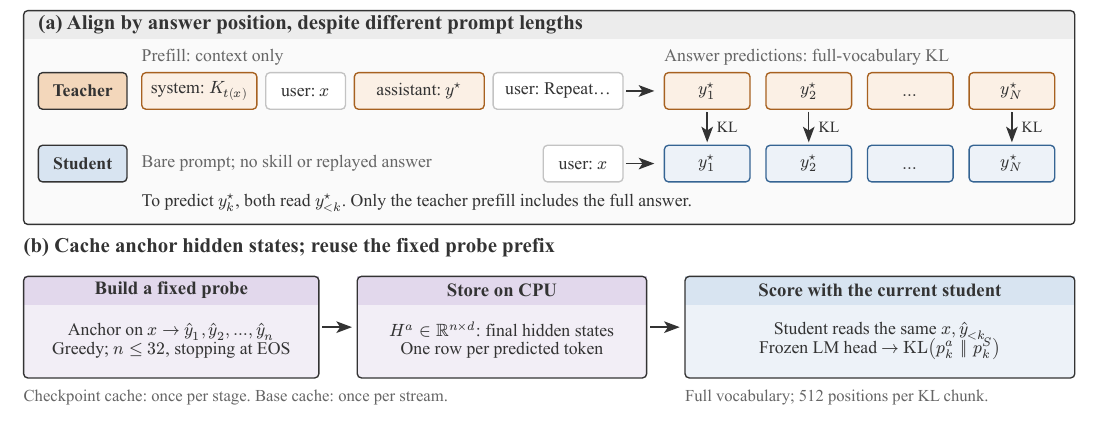}
    \appendixfigcaption
    \caption{Fixed-prefix scoring. (a) Vertical arrows compare teacher and student distributions at the same answer position by full-vocabulary KL. Both read $y^*_{<k}$; only the teacher's prefill contains the full answer. $N$ includes the closing end-of-turn token; chat-template delimiters are omitted. (b) Anchor hidden states are cached for a fixed greedy probe prefix and projected through the frozen output head when computing drift. Here $n\le32$ is the prefix length and $d$ the hidden width.}
    \label{fig:stage-ledger}
    \vspace{0.5\baselineskip}
}]

\section{Implementation Details}
\label{sec:appendix-implementation}

One model, Qwen3.5-4B \citep{qwen-team-2026-qwen35} at a pinned revision, fills every role; all calls go through its chat template with thinking disabled.
The weights and chat template are provided by the Qwen Team in the official model repository.\footnote{\url{https://huggingface.co/Qwen/Qwen3.5-4B}}
Figure~\ref{fig:stage-ledger} details answer-position alignment and probe caching.
A skill in the registry $K$ is an identifier, a routing prefix, and instructions.
A prompt is routed to the skill with the longest prefix that matches it after NFKC normalization, whitespace collapsing, and lowercasing; the instructions become a fixed system message, and a prompt with no match runs bare.
The serialized registry, prefixes included, counts toward $B_K=24{,}576$ tokens, and one skill holds at most 768.
The window has $B_A=8{,}192$ tokens; the memory has $B_R=32{,}768$, of which $B_G=8{,}192$ are reserved for $G$, so task records occupy at most 24,576.

At stage $t$, $A_t$ is filled with whole records of $D_t^{\mathrm{train}}$ in a seeded order fixed per stage, each counted as prompt plus answer plus a 16-token allowance, until the next record would overflow $B_A$.
The frozen $\bar\theta_{t-1}$ answers every record greedily; an answer that differs from $y^*$ after the same normalization is a mistake, and at most four reach the writer.
The writer is $\theta_0$ and decodes once, greedily, for at most 768 new tokens.
Its prompt lists up to 16 records of $A_t$ (those whose answers open differently first; inputs clipped to 384 tokens, outputs to 128), then the mistakes, and asks for a skill under four headers: Format, Rules, Avoid, Examples.
The skill's prefix is the longest common prefix of the task's training prompts; its instructions are the reply from its Format line on, clipped to 768 tokens; it replaces any skill with the same prefix or identifier.
The guard is the mean residual of the window under the frozen checkpoint,
\begin{equation}
\begin{gathered}
    \bar r_t(K)=\frac{1}{|A_t|}\sum_{(x,y^*)\in A_t} r(x,y^*;\bar\theta_{t-1},K),\\
    K_t=\begin{cases}
        K' & \begin{aligned}[t]
            &\text{if } \bar r_t(K')<\bar r_t(K_{t-1})\\
            &\text{and } \mathrm{size}(K')\le B_K,
        \end{aligned}\\
        K_{t-1} & \text{otherwise},
    \end{cases}
\end{gathered}
\label{eq:guard}
\end{equation}
so a candidate $K'$ is kept only when the skill alone, without a replayed answer in the prefill, makes the window easier to predict.

A residual (Eq.~\ref{eq:residual}) is scored with only the routed skill and bare prompt in the prefill, conditioning on $y^*_{<k}$ at each answer position.
The checkpoint probes of stage $t$ come from $\mathcal{R}_{t-1}$: up to eight prompts of at most 2,048 tokens per retained task, each completed by the first 32 greedy tokens of $\bar\theta_{t-1}$ on the bare prompt and grouped by task.
The general probes are the 48 records of $G$ within $B_G$, non-test GSM8K, IFEval, and Super-NaturalInstructions records disjoint from the G-test (Appendix~\ref{sec:appendix-experiments}), completed by the first 32 greedy tokens of $\theta_0$ once at the start of the stream and grouped by source.
Both probe types stop early at EOS and remain fixed during updates.
Each anchor's final hidden states at the probe's prediction positions are cached, once per stage for the checkpoint anchor and once per stream for the base anchor (Figure~\ref{fig:stage-ledger}b).
The frozen output head projects cached anchor states and current student states to full-vocabulary distributions; KL is computed in chunks of 512 positions.

A training record is one whose bare prompt and verified answer, the answer clipped to 16,384 tokens and closed by the end-of-turn token, fit 16,384 tokens together.
The teacher reads the routed skill as the system message, the prompt, the verified answer as its own previous turn, and a final user turn, ``Repeat your previous answer exactly, with no other text.''; its distribution at each answer position is computed with the adapter disabled.
Teacher and student both read the verified-answer prefix $y^*_{<k}$ at position $k$.
Their distributions are aligned by answer-token index rather than absolute sequence position, with the last prompt position predicting the first answer token (Figure~\ref{fig:stage-ledger}a).
The student adds LoRA factors of rank 32, $\alpha=64$, and no dropout to every linear layer of the text decoder, leaving out the vision tower, the multi-token-prediction head, and the output head.
A stage runs 32 updates of 32 records in a seeded shuffle, with AdamW without weight decay, a learning rate of $10^{-4}$ decaying linearly to zero, and gradient-norm clipping at 1.0; the final iterate is kept.
Each update first measures both drifts on all probes without gradient, then steps on
\begin{equation}
    \mathcal{L}(\phi)=\mathcal{L}_{\mathrm{dist}}(\phi)+\sum_{j\in\{C,0\}}[\lambda_j>0]\,\frac{\lambda_j}{\varepsilon_j}\,\bar D^{j}_{g_j^*}(\phi),
    \label{eq:update}
\end{equation}
where $[\cdot]$ is the indicator, $g_j^*$ is the worst group of drift $j$ at that measurement, and $\bar D^j_{g}$ is that group's mean divergence, the inner mean of Eq.~\ref{eq:drift}; it finally moves each multiplier by Eq.~\ref{eq:dual} from the same measurement, with $\varepsilon_C=\varepsilon_0=0.05$, $\eta=0.5$, and both multipliers starting at zero.
The drifts before the first update and after the last are recorded for every stage; the latter are the audited values of Figure~\ref{fig:long-horizon-appendix} and Table~\ref{tab:hero-ablation}.

The adapter is cast to bf16, saved, and merged into the dense weights, which become $\bar\theta_t$ and the backbone of the next stage's fresh adapter, so every stage starts from $\Delta W=0$.
The memory is then refreshed under the merged weights.
With $\rho_{(1)},\ldots,\rho_{(n)}$ the records of $\mathcal{R}_{t-1}\cup A_t$, deduplicated by identifier, in decreasing $r(\cdot;\bar\theta_t,K_t)$ with ties broken by identifier, $|\rho|$ a record's token length, and $\tau_i$ the token count of $\mathcal{R}^{(i)}$, one scan keeps
\begin{equation}
    \mathcal{R}^{(i)}=\begin{cases}
        \mathcal{R}^{(i-1)}\cup\{\rho_{(i)}\} & \begin{aligned}[t]
            &\text{if } \tau_{i-1}{+}|\rho_{(i)}|\\
            &\qquad\le B_R{-}B_G,
        \end{aligned}\\
        \mathcal{R}^{(i-1)} & \text{otherwise},
    \end{cases}
    \label{eq:memory}
\end{equation}
from $\mathcal{R}^{(0)}=\emptyset$, and $\mathcal{R}_t=\mathcal{R}^{(n)}$: a record enters only if it fits whole, and a smaller record further down the order may still enter after a larger one is skipped.
Within a stage, one copy of $\bar\theta_{t-1}$ with its adapter serves as student, teacher, and checkpoint anchor; $\theta_0$ is loaded only to write the skill.
Training uses PyTorch 2.6.0 \citep{paszke-etal-2019-pytorch}, Transformers 5.2.0 \citep{wolf-etal-2020-transformers}, and PEFT 0.20.0 \citep{mangrulkar-etal-2022-peft} in bf16.
Each stage records its registry, guard decision, memory, drifts, and the adapter's SHA-256, from which a restarted job replays completed stages.

\section{Experiment Details}
\label{sec:appendix-experiments}

TRACE+17 is built once, with seed 42, from pinned revisions of the TRACE and Super-NaturalInstructions repositories: the eight TRACE tasks in their canonical order, then SuperNI tasks 024, 074, 195, 511, 1219, 574, 192, 177, 064, 346, 108, 111, 1415, 1447, 828, 1728, and 217.
Prompts are deduplicated and the splits drawn in a seeded hash order.
Each task keeps its official metric: exact match, leading-choice accuracy, ROUGE-L \citep{lin-2004-rouge}, edit similarity, or SARI \citep{xu-etal-2016-optimizing}.
Validation targets and test records are hidden from the model, and no test score selects a candidate or hyperparameter.
The G-test draws from pinned snapshots of MMLU, GSM8K, HellaSwag, IFEval, and XCOPA, disjoint from the stream: 200 test items per category, a 40-item probe nested in them, and a separate 40-item development panel.
Stage 0 is the base model.
After stage $t$ we score the current task; at stages 4, 8, 16, and 25 every seen task and the full G-test; at stages 12 and 20 tasks 1 to 8; at every other stage the 200-item probe.
Each cell uses all 100 test items, decoded greedily with thinking off, at most 5,120 new tokens, and 16,384 input tokens; an input over the ceiling counts as an empty prediction.
LoRA baselines decode merged checkpoints with vLLM 0.31.0 \citep{kwon-etal-2023-pagedattention}; HERD and MoRAM decode with Transformers; frozen-model methods use MLX \citep{hannun-etal-2023-mlx}, PyTorch's MPS backend,\footnote{\url{https://docs.pytorch.org/docs/stable/notes/mps.html}} or vLLM servers.
GSM8K is scored on the final number, IFEval by its strict verifier, and the choice categories by one extractor that reads the last explicit answer.

Every weight-updating method trains for 32 updates of 32 records per stage on sequences of at most 16,384 tokens; longer records are excluded, not truncated.
Unless the official code fixes otherwise, the adapter is a rank-32 LoRA with $\alpha=64$ and no dropout on every linear layer of the text backbone, trained with AdamW at $10^{-4}$ and gradient clipping at 1.0.
Uniform replay keeps a 32,768-token memory, 8,192 of them for 23 reserved general records, filled by uniform sampling without re-entry.
OPR-RU samples one response per earlier training prompt at temperature 0.1 and keeps the best-scoring five per stage (replay ratio 0.01), with an equal quota per task.
TreeLoRA uses the authors' regularizer weight of 0.5.
MoRAM follows the official runner: eight rank-1 atoms per module per stage, $\alpha=16$, dropout 0.1, top-8 self-activation at temperature 0.03 with an inference gate of 0.2, and Adam$(0.9,0.95)$ at $5\times10^{-4}$.
SDFT trains a fresh adapter per stage on 32 of its own rollouts per update, at temperature 1 and at most 1,024 new tokens; its teacher is an EMA copy (0.01) of the stage-start weights shown the reference under the official thinking-process template.
OPHSD uses the official draft-verify harness on the six stages with at most ten labels and plan-solve elsewhere; the teacher is the stage-start model given the draft labels, or the plan and the reference answer, and the student minimizes the reverse KL on one on-policy rollout per record.

The static harness wraps the frozen model in a fixed system prompt, an empty skill registry, and a Python tool in a 10-second sandbox, for at most 32 turns per item.
ACE runs the official offline trainer: batch size 1, one epoch, three reflection rounds, ADD-only curation, the upstream 4,096-token call limit, and a rollback of any sample whose call exceeds the input ceiling.
HCL-Stable proposes one alternative per component per stage under a 12,000-token budget and commits it only on a validation gain of at least 0.01 that breaks no item of each earlier task's 20-item validation panel.
Learning methods run with seeds 42, 43, and 44 and report the mean and sample standard deviation; intervals are 95\% item bootstraps with 4,000 draws, resampled within task and G-test category and paired between HERD and each baseline.

\paragraph{Original dataset sources.}
In addition to the TRACE benchmark citation, we credit the creators of its constituent datasets: C-STANCE \citep{zhao-etal-2023-c}, FOMC \citep{shah-etal-2023-trillion}, MeetingBank \citep{hu-etal-2023-meetingbank}, Py150 \citep{raychev2016probabilistic} as used in CodeXGLUE \citep{lu-etal-2021-codexglue}, ScienceQA \citep{lu-etal-2022-scienceqa}, the two NumGLUE tasks \citep{mishra-etal-2022-numglue}, and 20Minuten \citep{rios-etal-2021-new}.

The 17 additional tasks use the instruction-based formulations from Super-NaturalInstructions \citep{wang-etal-2022-super}; the task identifiers map to the following original resources:
024, Cosmos QA \citep{huang-etal-2019-cosmos};
074, SQuAD 1.1 \citep{rajpurkar-etal-2016-squad};
195, Sentiment140 \citep{go-etal-2009-twitter};
511, Reddit TIFU \citep{kim-etal-2019-abstractive};
1219, the multilingual TED corpus, English-to-Spanish translation \citep{qi-etal-2018-pre};
574, AirDialogue \citep{wei-etal-2018-airdialogue};
192, HotpotQA \citep{yang-etal-2018-hotpotqa};
177, ParaNMT \citep{wieting-gimpel-2018-paranmt};
064, the synthetic list-processing task contributed by Pulkit Verma to Super-NaturalInstructions;\footnote{\url{https://github.com/allenai/natural-instructions/blob/master/tasks/task064_all_elements_except_first_i.json}}
346, HybridQA \citep{chen-etal-2020-hybridqa};
108, the Contextual Abuse Dataset \citep{vidgen-etal-2021-introducing};
111, ASSET \citep{alva-manchego-etal-2020-asset};
1415, Emily McMilin's YouTube Caption Corrections;\footnote{\url{https://github.com/2dot71mily/youtube_captions_corrections}}
1447, ADE Corpus V2 \citep{gurulingappa-etal-2012-ade};
828, COPA \citep{roemmele-etal-2011-copa};
1728, WebNLG \citep{gardent-etal-2017-webnlg};
and 217, ROCStories \citep{mostafazadeh-etal-2016-corpus}.
The Super-NaturalInstructions task files identify both the original sources and the contributors of the instruction-based formulations.\footnote{\url{https://github.com/allenai/natural-instructions/tree/master/tasks}}

\section{Additional Experimental Results}
\label{sec:appendix-results}
\setcounter{dbltopnumber}{2}
\renewcommand{\dbltopfraction}{0.9}
\renewcommand{\textfraction}{0.08}
\renewcommand{\dblfloatpagefraction}{0.75}

\begin{figure*}[t]
    \centering
    \includegraphics[width=\textwidth]{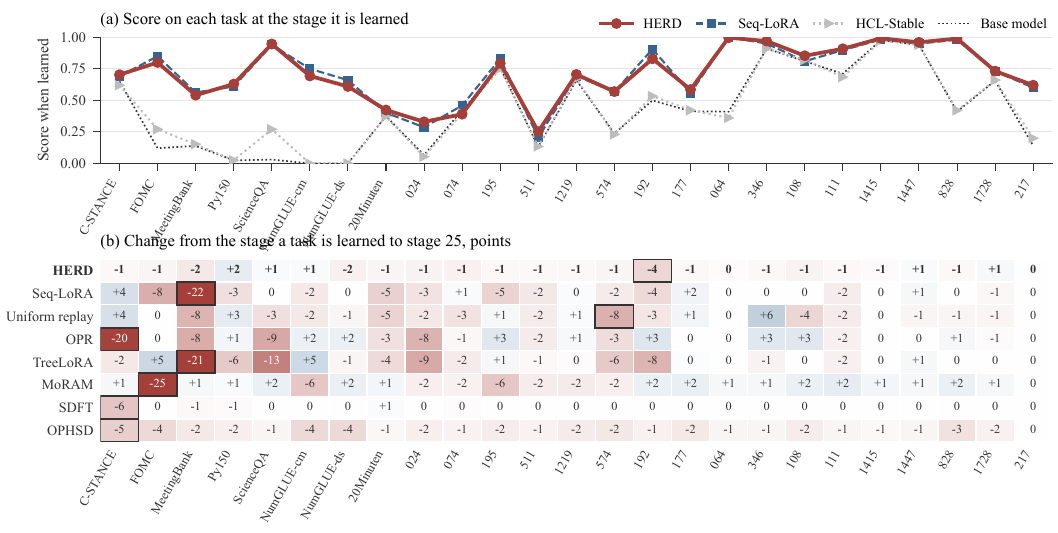}
    \caption{Table~\ref{tab:main-results} by task; means over three seeds. (a) Score on each task at the stage it is learned, for HERD, Seq-LoRA, HCL-Stable, and the base model. (b) Change on each task between the stage it is learned and stage 25, in points, for the weight-updating methods; the outlined cell is each method's largest loss.}
    \label{fig:task-matrix}
\end{figure*}

\paragraph{Acquisition and retention by task.}
Figure~\ref{fig:task-matrix} splits Table~\ref{tab:main-results} into its 25 tasks.
At the stage a task is learned (panel a), HERD and Seq-LoRA move together, within 3 points on 16 tasks and never more than 7.7 apart (SuperNI 192); HERD is 60 to 92 points above the base on FOMC, Py150, ScienceQA, and the two NumGLUE tasks, on which the base scores at most 0.12.
HCL-Stable, which commits nothing, follows the base line.
Between that stage and stage 25 (panel b), HERD moves by at most 4 points on any task.
Seq-LoRA, TreeLoRA, OPR, and MoRAM each lose 20 points or more on one task, and not the same one: Seq-LoRA and TreeLoRA on MeetingBank (22 and 21), MoRAM on FOMC (25), OPR on C-STANCE (20); uniform replay's largest losses are 8 points, on MeetingBank and SuperNI 574, and OPHSD loses 1 to 5 points on every earlier task.
In the parametric baselines, forgetting is concentrated on a few tasks rather than spread over all of them, which is what the worst-group drift of Eq.~\ref{eq:drift} is defined to catch and a mean over probes would miss.

\begin{table*}[t]
    \centering
    \resizebox{\textwidth}{!}{\footnotesize%
\setlength{\tabcolsep}{4pt}
\begin{tabular}{lccccc}
\toprule
Baseline & $\Delta$AA & $\Delta$Acq. & $\Delta$F & $\Delta$BWT & $\Delta$G-test \\
\midrule
Base & \textbf{$+$0.287\,{\scriptsize[0.274, 0.299]}} & \textbf{$+$0.293\,{\scriptsize[0.281, 0.306]}} & \textbf{$+$0.010\,{\scriptsize[0.006, 0.016]}} & \textbf{$-$0.007\,{\scriptsize[$-$0.013, $-$0.001]}} & \textbf{$-$0.063\,{\scriptsize[$-$0.088, $-$0.038]}} \\
Static harness & \textbf{$+$0.274\,{\scriptsize[0.262, 0.286]}} & \textbf{$+$0.281\,{\scriptsize[0.268, 0.293]}} & \textbf{$+$0.010\,{\scriptsize[0.006, 0.016]}} & \textbf{$-$0.007\,{\scriptsize[$-$0.013, $-$0.002]}} & $-$0.017\,{\scriptsize[$-$0.044, 0.010]} \\
\midrule
HCL-Stable & \textbf{$+$0.273\,{\scriptsize[0.260, 0.286]}} & \textbf{$+$0.278\,{\scriptsize[0.265, 0.291]}} & \textbf{$+$0.007\,{\scriptsize[0.003, 0.013]}} & $-$0.005\,{\scriptsize[$-$0.012, 0.001]} & \textbf{$-$0.049\,{\scriptsize[$-$0.074, $-$0.024]}} \\
ACE & \textbf{$+$0.279\,{\scriptsize[0.266, 0.292]}} & \textbf{$+$0.285\,{\scriptsize[0.272, 0.298]}} & \textbf{$+$0.009\,{\scriptsize[0.005, 0.015]}} & $-$0.006\,{\scriptsize[$-$0.013, 0.000]} & \textbf{$+$0.026\,{\scriptsize[0.001, 0.051]}} \\
\midrule
Seq-LoRA & \textbf{$+$0.012\,{\scriptsize[0.004, 0.021]}} & $-$0.002\,{\scriptsize[$-$0.010, 0.007]} & \textbf{$-$0.015\,{\scriptsize[$-$0.024, $-$0.009]}} & \textbf{$+$0.015\,{\scriptsize[0.006, 0.023]}} & \textbf{$+$0.177\,{\scriptsize[0.154, 0.200]}} \\
Uniform replay & $+$0.007\,{\scriptsize[$-$0.001, 0.016]} & $+$0.001\,{\scriptsize[$-$0.007, 0.010]} & \textbf{$-$0.010\,{\scriptsize[$-$0.020, $-$0.004]}} & $+$0.006\,{\scriptsize[$-$0.004, 0.017]} & \textbf{$+$0.175\,{\scriptsize[0.150, 0.201]}} \\
OPR & \textbf{$+$0.016\,{\scriptsize[0.007, 0.025]}} & \textbf{$+$0.008\,{\scriptsize[0.000, 0.015]}} & \textbf{$-$0.014\,{\scriptsize[$-$0.023, $-$0.007]}} & $+$0.009\,{\scriptsize[$-$0.001, 0.018]} & \textbf{$+$0.160\,{\scriptsize[0.136, 0.184]}} \\
TreeLoRA & \textbf{$+$0.020\,{\scriptsize[0.011, 0.029]}} & $+$0.002\,{\scriptsize[$-$0.006, 0.009]} & \textbf{$-$0.021\,{\scriptsize[$-$0.030, $-$0.015]}} & \textbf{$+$0.019\,{\scriptsize[0.010, 0.029]}} & \textbf{$+$0.169\,{\scriptsize[0.144, 0.194]}} \\
MoRAM & \textbf{$+$0.013\,{\scriptsize[0.005, 0.022]}} & \textbf{$+$0.012\,{\scriptsize[0.004, 0.020]}} & \textbf{$-$0.009\,{\scriptsize[$-$0.018, $-$0.002]}} & $+$0.002\,{\scriptsize[$-$0.007, 0.011]} & \textbf{$+$0.137\,{\scriptsize[0.114, 0.161]}} \\
\midrule
OPHSD & \textbf{$+$0.087\,{\scriptsize[0.073, 0.100]}} & \textbf{$+$0.074\,{\scriptsize[0.061, 0.087]}} & \textbf{$-$0.010\,{\scriptsize[$-$0.017, $-$0.005]}} & \textbf{$+$0.014\,{\scriptsize[0.007, 0.020]}} & \textbf{$+$0.192\,{\scriptsize[0.166, 0.218]}} \\
\bottomrule
\end{tabular}
}
    \caption{HERD minus each baseline after stage 25 on the quantities of Table~\ref{tab:main-results}, with 95\% paired item-bootstrap intervals (4,000 draws, items resampled within task and G-test category). Bold: the interval excludes zero.}
    \label{tab:comparisons}
\end{table*}

\paragraph{Paired comparisons.}
Table~\ref{tab:comparisons} lists the differences behind \S\ref{sec:experiment-results}; each ``significantly'' there is an interval here that excludes zero.
SDFT is omitted: at AA 0.021 it has collapsed (\S\ref{sec:experiment-results}), and an interval on its difference would test nothing about retention.
The intervals that include zero are the ties of \S\ref{sec:experiment-results}: acquisition against Seq-LoRA, uniform replay, and TreeLoRA, AA against uniform replay, and BWT against half the baselines.
The G-test differences to the parametric baselines in the table, 13.7 to 17.7 points, have intervals no closer than 11.4 points to zero.
Against the static harness the G-test difference is $-1.7$ points with an interval that includes zero: after 25 stages of training, HERD's general capability is statistically indistinguishable from the frozen base under a fixed system prompt.

\begin{figure*}[t]
    \centering
    \includegraphics[width=\textwidth]{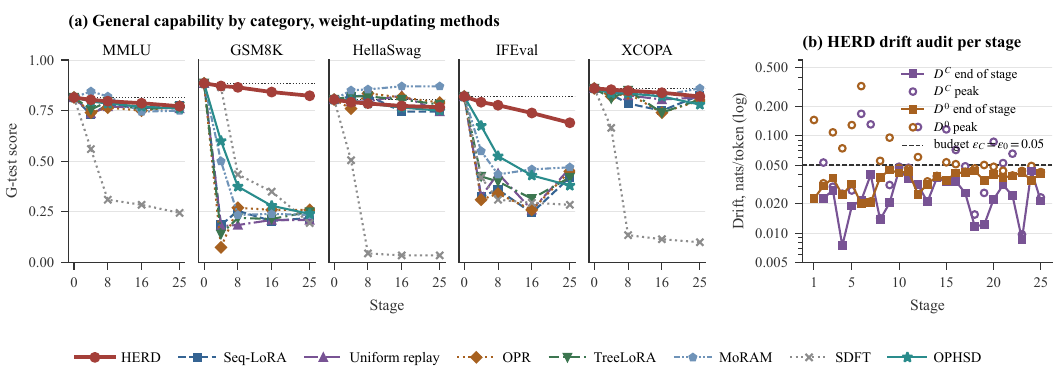}
    \caption{Per-stage view behind Figure~\ref{fig:experimental-summary}. (a) G-test score per category at stages 0, 4, 8, 16, and 25 for the weight-updating methods, starting from the base model (dotted). (b) HERD's audited checkpoint drift $D^C$ and base drift $D^0$ at the end of each stage (filled) and their within-stage peaks (hollow) against the budgets (dashed), on a log axis.}
    \label{fig:long-horizon-appendix}
\end{figure*}

\paragraph{General capability by category.}
Figure~\ref{fig:long-horizon-appendix}a is the per-category view behind Figure~\ref{fig:experimental-summary}c.
The parametric baselines lose GSM8K and IFEval between stage 0 and stage 4, Seq-LoRA by 70 and 49 points, and stay there; MMLU, HellaSwag, and XCOPA move by at most 12 points for every method but SDFT, whose five curves fall together.
HERD's categories are within 3 points of the base at stage 4 and within 6 at stage 25, except instruction following, which loses 13 points, 9 of them after stage 8.
The 200-item probe scored at the other stages shows what the anchors skip: Seq-LoRA's probe mean falls to 0.28, 0.16, and 0.37 at stages 11 to 13 and is back at 0.59 by stage 14, a collapse and recovery that lies between the anchors of Figure~\ref{fig:experimental-summary}b; HERD's probe mean never falls below 0.75.

\begin{figure*}[t]
    \centering
    \includegraphics[width=\textwidth]{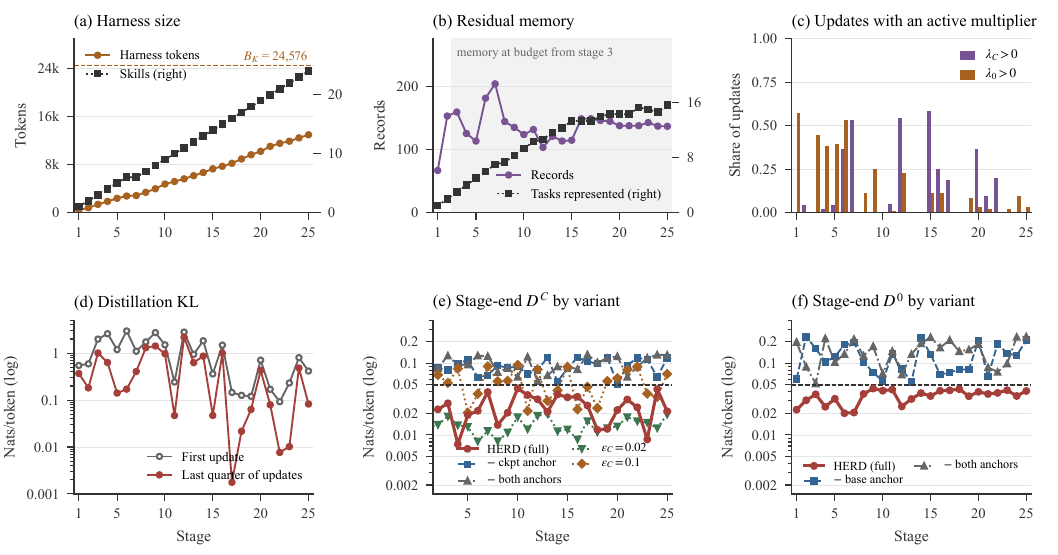}
    \caption{HERD's state by stage; means over three seeds. (a) Registry size in tokens against $B_K$, and skills in the registry. (b) Residual memory: whole records held and tasks represented; in the shaded stages the memory is at its token budget. (c) Share of a stage's 32 updates at which each multiplier is positive. (d) Distillation KL (Eq.~\ref{eq:distill}) at the first update and averaged over the last eight. (e, f) Audited stage-end $D^C$ and $D^0$ for full HERD and the single-seed variants of Table~\ref{tab:hero-ablation} that change them; dashed: the 0.05 budget.}
    \label{fig:mechanism}
\end{figure*}

\paragraph{Drift audit.}
Figure~\ref{fig:long-horizon-appendix}b audits HERD's two drifts at every stage.
The three-seed mean stage-end values never exceed the budget: $D^C$ reaches at most 0.045 and $D^0$ at most 0.044.
Within a stage, the highest measurement exceeds 0.05 at 9 stages for $D^C$ and 11 for $D^0$, up to 0.168 and 0.324 at stage 6, and is pulled back before the stage ends; after stage 9 the $D^0$ peak is at most 0.060.
Figure~\ref{fig:mechanism}e,f put the variants of Table~\ref{tab:hero-ablation} on the same axes.
Without the checkpoint anchor, $D^C$ ends above its budget at all 25 stages, between 0.052 and 0.121.
Without the base anchor, $D^0$ ends between 0.055 and 0.235, and the checkpoint constraint that is kept also ends over its budget at 15 stages.
At $\varepsilon_C=0.02$ the stage-end $D^C$ stays between 0.008 and 0.020; at $\varepsilon_C=0.1$ it stays under 0.1 and is over 0.05 at 15 stages.

\begin{table*}[t]
    \centering
    {\small%
\setlength{\tabcolsep}{4pt}
\begin{tabular}{lccccccc}
\toprule
Variant & Acq.$\uparrow$ & AA$\uparrow$ & F$\downarrow$ & $\Delta C\uparrow$ & $\Delta C_{\min}\uparrow$ & $D^C$ & $D^0$ \\
\midrule
\textbf{HERD (full)} & 0.713{\scriptsize$\pm$0.002} & 0.706{\scriptsize$\pm$0.008} & 0.010{\scriptsize$\pm$0.003} & $-$0.063{\scriptsize$\pm$0.006} & $-$0.130{\scriptsize$\pm$0.012} & 0.021 & 0.041 \\
\midrule
$-$ skills\textsuperscript{*} & \textbf{0.668\,{\scriptsize($-$0.045)}} & \textbf{0.656\,{\scriptsize($-$0.050)}} & 0.013\,{\scriptsize($+$0.003)} & $-$0.073\,{\scriptsize($-$0.010)} & $-$0.135\,{\scriptsize($-$0.005)} & 0.021 & 0.040 \\
uniform memory\textsuperscript{*} & 0.708\,{\scriptsize($-$0.005)} & 0.690\,{\scriptsize($-$0.017)} & 0.020\,{\scriptsize($+$0.010)} & $-$0.071\,{\scriptsize($-$0.008)} & $-$0.125\,{\scriptsize($+$0.005)} & 0.034 & 0.041 \\
$-$ checkpoint anchor\textsuperscript{*} & 0.716\,{\scriptsize($+$0.004)} & 0.688\,{\scriptsize($-$0.018)} & 0.030\,{\scriptsize($+$0.020)} & $-$0.075\,{\scriptsize($-$0.012)} & $-$0.140\,{\scriptsize($-$0.010)} & \textbf{0.118} & 0.044 \\
$-$ base anchor\textsuperscript{*} & 0.718\,{\scriptsize($+$0.006)} & 0.704\,{\scriptsize($-$0.002)} & 0.016\,{\scriptsize($+$0.006)} & $-$0.158\,{\scriptsize($-$0.095)} & $-$0.490\,{\scriptsize($-$0.360)} & 0.027 & \textbf{0.212} \\
$-$ both anchors\textsuperscript{*} & 0.721\,{\scriptsize($+$0.008)} & 0.680\,{\scriptsize($-$0.026)} & \textbf{0.044\,{\scriptsize($+$0.033)}} & \textbf{$-$0.193\,{\scriptsize($-$0.130)}} & \textbf{$-$0.560\,{\scriptsize($-$0.430)}} & \textbf{0.131} & \textbf{0.236} \\
\midrule
SFT objective\textsuperscript{*} & 0.722\,{\scriptsize($+$0.009)} & 0.706\,{\scriptsize($-$0.001)} & 0.018\,{\scriptsize($+$0.008)} & $-$0.123\,{\scriptsize($-$0.060)} & $-$0.320\,{\scriptsize($-$0.190)} & 0.047 & 0.048 \\
$\varepsilon_C$ $=$ 0.02\textsuperscript{*} & 0.695\,{\scriptsize($-$0.018)} & 0.691\,{\scriptsize($-$0.015)} & 0.005\,{\scriptsize($-$0.005)} & $-$0.051\,{\scriptsize($+$0.012)} & $-$0.085\,{\scriptsize($+$0.045)} & 0.019 & 0.038 \\
$\varepsilon_C$ $=$ 0.1\textsuperscript{*} & 0.717\,{\scriptsize($+$0.004)} & 0.699\,{\scriptsize($-$0.008)} & 0.021\,{\scriptsize($+$0.011)} & $-$0.083\,{\scriptsize($-$0.020)} & $-$0.160\,{\scriptsize($-$0.030)} & 0.071 & 0.046 \\
\midrule
+ harness (infer.)\textsuperscript{\ddag} & 0.713\,{\scriptsize(0.000)} & 0.708\,{\scriptsize($+$0.002)} & 0.007\,{\scriptsize($-$0.003)} & $-$0.061\,{\scriptsize($+$0.002)} & $-$0.128\,{\scriptsize($+$0.002)} & 0.021 & 0.041 \\
\bottomrule
\end{tabular}
\par\smallskip{\scriptsize \textsuperscript{\ddag}\,inference only: the deployed weights of the full run re-scored with the stage-25 harness attached to every prompt. \textsuperscript{*}\,single seed.}
}
    \caption{HERD ablations on TRACE+17, the full table behind Figure~\ref{fig:hero-ablation}: scores after stage 25 with the difference from full HERD in parentheses; $D^C$ and $D^0$ are the audited stage-25 drifts against the budgets $\varepsilon_C=\varepsilon_0=0.05$. Bold marks the variant that hurts most in each column and any drift over its budget.}
    \label{tab:hero-ablation}
\end{table*}

\paragraph{Harness, memory, and multipliers by stage.}
Figure~\ref{fig:mechanism}a--d follow HERD's state through the stream.
The registry gains one skill per task, 24 in all because the stage-7 skill replaced the stage-6 skill under the shared NumGLUE prefix, and ends at 12,982 tokens, 53\% of $B_K$.
The memory is at its budget from stage 3 on, within 150 tokens of 24,576; it holds 103 to 204 whole records, 137 at stage 25 from about 16 of the 25 tasks seen.
The base multiplier is active at 57\% of the updates of stage 1 and at no more than 12\% after stage 12; the checkpoint multiplier is never active at 12 stages and at most at 58\% of the updates of one stage.
The distillation KL falls within every stage.
At the first update it is largest on the tasks the base model answers worst, above 2.5 nats per token on Py150, NumGLUE-cm, and SuperNI 024 and 511, and below 0.25 on six of the last nine SuperNI tasks.

\begin{table}[t]
    \centering
    {\footnotesize%
\setlength{\tabcolsep}{4pt}
\begin{tabular}{lrrrr}
\toprule
Method & \multicolumn{3}{c}{Training time (s)} & Context \\
\cmidrule(lr){2-4}
 & stage 1 & stage 25 & total & tokens \\
\midrule
\textbf{HERD} & 316 & 284 & 10,723 & 0 \\
\midrule
HCL-Stable & 580 & 296 & 46,887 & 39 \\
ACE & 60,412 & 36,759 & 888,082 & 15,410 \\
\midrule
Seq-LoRA & 51 & 200 & 10,463 & 0 \\
Uniform replay & 318 & 400 & 15,267 & 0 \\
OPR & 667 & 705 & 18,670 & 0 \\
TreeLoRA & 157 & 256 & 12,733 & 0 \\
MoRAM & 171 & 433 & 14,017 & 0 \\
SDFT & 1,999 & 3,626 & 122,508 & 0 \\
\midrule
OPHSD & 11,364 & 30,495 & 659,000 & 0 \\
\bottomrule
\end{tabular}
}
    \caption{Training time per stage and in total, and the context attached to every prompt at inference.}
    \label{tab:cost}
\end{table}

\paragraph{Cost by stage.}
Table~\ref{tab:cost} gives the training time of every learning method at stage 1, at stage 25, and in total.
HERD's stage time follows the task's answer length and not the stage index, 316~s at stage 1 and 284~s at stage 25, as does OPR's, 667 and 705~s; HERD's total of 10,723~s is within 3\% of Seq-LoRA's.
The two recursive self-distillation baselines grow with the stage, SDFT from 1,999 to 3,626~s as its rollouts lengthen (\S\ref{sec:experiment-results}) and OPHSD from 11,364 to 30,495~s as it regenerates its teacher context by rollouts.
ACE's 888,082~s go to the 38,103 generator, reflector, and curator calls of its offline trainer, and its deployed playbook adds 15,410 tokens to every prompt; HERD deploys bare weights.

\paragraph{Ablations.}
\label{sec:appendix-ablation}
\label{sec:appendix-paired}
Table~\ref{tab:hero-ablation} is the full table behind Figure~\ref{fig:hero-ablation}, with the SFT objective, the $\varepsilon_C$ sweep, and the deployed weights re-scored with the harness.
At $\varepsilon_C=0.02$ the constraint binds: acquisition pays 1.8 points and $\Delta C_{\min}$ improves by 4.5.
At $\varepsilon_C=0.1$ it goes slack: forgetting rises by 1.1 points and the audited $D^C$ ends at 0.071, over 0.05 and under the budget it was given.
The smaller moves in the other columns, 0.5 points of forgetting at 0.02 and 0.4 of acquisition at 0.1, are inside the seed spread.
$\varepsilon_0$ was not swept: removing the base anchor is its $\varepsilon_0\to\infty$ limit and already costs 9.5 points of $\Delta C$.
The last row scores the stage-25 weights with the stage-25 harness prepended to every prompt: AA moves from 0.706 to 0.708, F from 0.010 to 0.007, each G-test category by under 0.2 point, and no task by more than 0.4 points.
The results indicate that the model has already internalized the harness.

\paragraph{Seeds and uncertainty.}
Every number in Tables~\ref{tab:main-results} and \ref{tab:comparisons} and in Figures~\ref{fig:experimental-summary}, \ref{fig:long-horizon-appendix}, and \ref{fig:task-matrix} is a mean over seeds 42, 43, and 44; the task and G-test cells carry their sample standard deviation at every anchor stage, and every interval is a 95\% item bootstrap with 4,000 draws.
The ablations and the $\varepsilon_C$ sweep are single runs, because the claim is the ordering of the variants and not their level.
Three times the full model's displayed seed standard deviations gives approximate reference scales of 0.006 for acquisition, 0.024 for AA, 0.009 for F, 0.018 for $\Delta C$, and 0.036 for $\Delta C_{\min}$.
Gaps within that spread, such as the 0.4-point acquisition change without the checkpoint anchor, are reported in Table~\ref{tab:hero-ablation} and not interpreted.

\section{Case Studies}
\label{sec:appendix-cases}

\begin{figure*}[t]
    \centering
    \includegraphics[width=\textwidth]{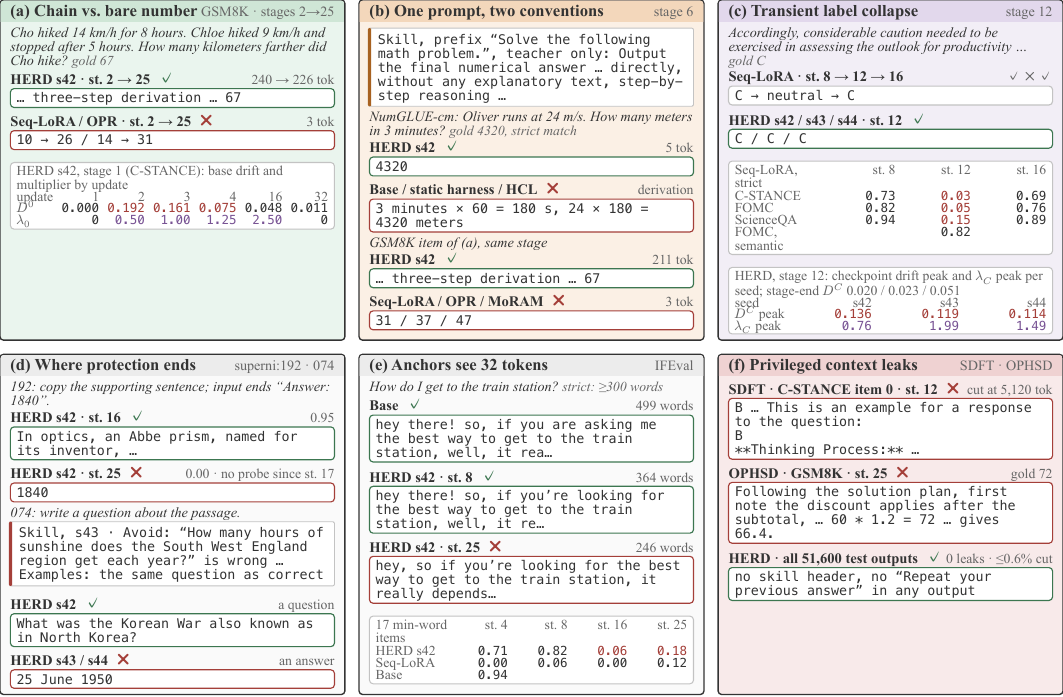}
    \caption{Single test items through the stream; every quoted output is copied from the run records, seed 42 unless marked. (a--c) HERD against the parametric baselines: a GSM8K chain kept through the one-letter stages, with stage 1's base drift and multiplier by update; a bare number on NumGLUE-cm beside a chain on GSM8K at stage 6; Seq-LoRA's label collapse at stage 12, with HERD's checkpoint-drift peaks. (d, e) Where HERD's protection ends: a learned task that lost its probes, a self-contradictory skill, and answer length beyond the 32-token anchor. (f) The privileged teacher context in SDFT's and OPHSD's outputs, and HERD's leak count.}
    \label{fig:cases}
\end{figure*}

\paragraph{HERD against the parametric baselines.}
Figure~\ref{fig:cases} follows single test items through the stream.
The first two tasks, C-STANCE and FOMC, train on one-letter answers, and 32 cross-entropy updates on them make the short answer a habit: from stage 2 on, the four parametric baselines answer GSM8K with a bare number, median 3 tokens.
On the Cho and Chloe item (Figure~\ref{fig:cases}a), the base and HERD write the same three-step derivation to 67 at stages 2 and 25, while Seq-LoRA, uniform replay, OPR, and TreeLoRA guess 10, 11, 14, and 14, then 26, 13, 31, and 22.
The behaviours part at the first tokens, where the base anchor looks: in stage 1 of seed 42, one update lifts $D^0$ to 0.192, $\lambda_0$ climbs to 2.5, and the drift is back under 0.05 before the stage ends.
At stage 6 (Figure~\ref{fig:cases}b), NumGLUE-cm prompts differ from GSM8K's by one word and demand a bare number.
HERD's stage-6 skill says so and is routed to the NumGLUE prefix alone: HERD answers 4320 on the Oliver item and keeps a 211-token chain on the GSM8K item, every parametric baseline answers both with a number, and the base, the static harness, and HCL derive 4320 inside a sentence and score zero.

\paragraph{Retention between the checkpoints.}
Some losses never reach Table~\ref{tab:main-results}.
After stage 12, superni:511 with free-text summaries, Seq-LoRA's C-STANCE, FOMC, and ScienceQA fall to 0.03, 0.05, and 0.15 and are back at 0.69, 0.76, and 0.89 by stage 16 (Figure~\ref{fig:cases}c).
The knowledge is intact: it answers FOMC with ``dovish'' or ``hawkish'' instead of a letter, and mapping the words back gives 0.82, its stage-8 score.
HERD's checkpoint probes open with the letter itself, so the same shift registers as $D^C$: the three seeds peak at 0.114 to 0.136, $\lambda_C$ rises to 0.76 to 1.99, the stage ends at 0.020 to 0.051, and FOMC stays at 0.75 to 0.80.
On MeetingBank, learned at stage 3, HERD's three seeds still write the ordinance title of item 65 at stage 25 (ROUGE-L 0.73) where Seq-LoRA writes ``The bill was passed.'' (0.00), without replaying any earlier answers.

\paragraph{HERD's failing cases.}
The first limitation is probe coverage.
Protection follows the residual memory (Figure~\ref{fig:cases}d,e): a learned task has low residual, leaves the memory, and loses its checkpoint probes with it.
Superni:192, copying the sentence that supports a given answer, lost its last probe at stage 17 in all three seeds; seed 42 then fell from 0.83 to 0.52 by stage 25, 70 of its 100 outputs copying the answer already in the input, while seeds 43 and 44 held above 0.90.
At stage 25, nine or ten of the 24 earlier tasks have no probe; a floor of probes per learned task is the obvious repair.
The second limitation is skill quality.
For superni:074, writing a question about a passage, seed 42's skill demands a question and 89 of 100 outputs are questions; the skills of seeds 43 and 44 list a question under Avoid that recurs under Examples, and only 21 and 64 outputs are questions, for ROUGE-L of 0.326 and 0.349 against the base's 0.421.
The guard passed all three, as it passed all 75 commits.
It checks the window's mean residual without replaying the answer (Eq.~\ref{eq:guard}); improvement under verified-answer prefixes does not guarantee that free generation follows the skill's intended format.
The third limitation is probe length.
Anchors compare 32 tokens, so on IFEval items with a minimum word count HERD opens like the base, then stops at 232 to 258 words against the base's 499; its pass rate on these 17 items falls from 0.71 to 0.76 at stage 4 to 0.18 to 0.24 at stage 25, against the base's 0.94.
Finally, the budget is a mean over a group's probes and tokens, so a few prompts can flip while the stage-end $D^0$ stays under 0.05: by stage 25, seeds 42 and 43 answer 44 and 60 of the 200 GSM8K items with a bare number, seed 44 only 6, at an accuracy still of 0.81 to 0.83.

\paragraph{OPHSD: drift in wording.}
OPHSD's outputs stay clean to the end, at most 1.3\% cut off, and its gains fall on the tasks the base gets right in the wrong format, ScienceQA from 0.03 to 0.85 and NumGLUE from 0 to about 0.5.
The official solve message gives the teacher a plan and a ``Reference answer (for verification only)'', and the student, whose prompt holds neither, begins to cite them (Figure~\ref{fig:cases}f): the share of test outputs that mention a plan or reference answer rises from 0.01 at stage 1 to 0.17 at stage 25, and to 0.84 on the mathematics items, whose outputs shorten from about 1,180 to 405 tokens.
Mathematics ends at 0.24 and instruction following at 0.38, against the base's 0.89 and 0.82, with planning-style openers in about 37\% of answers against 4\% for the base.
The harness is run by the current student and the teacher is the stage-start model, so nothing pulls back: the general probe never rises at any of the 25 stages.

\paragraph{SDFT: collapse by leakage and length.}
SDFT's teacher prompt shows an example and asks for ``a response of your own, including the thinking process''.
Here the example is a one- to three-token label, so the teacher invents a thinking process, and rollouts cut at 1,024 tokens are trained on regardless, so stopping is never supervised.
The drift compounds: test outputs cut at the 5,120-token ceiling go from 0.01 at stage 1 to 0.21, 0.74, and 0.94 at stages 4, 8, and 12, and the share that reproduces the teacher's instruction verbatim from 0.00 to 0.23, 0.77, and 0.95.
On C-STANCE item 0, stage 12 gives the letter, then ``This is an example for a response to the question:'' and a new thinking process, restarted until the ceiling (Figure~\ref{fig:cases}f).
The loss gives no warning, settling at 0.005 to 0.008, because the teacher is an EMA of the student.
HERD's target is the verified answer replayed as the model's own turn, so there is no instruction text to copy, and its KL is scored on that answer's prefix rather than on the student's samples, where teacher and student can agree on a degenerate path.

\section{Future Works}
\label{sec:appendix-future-work}

\paragraph{Larger streams.}
Our experiments cover one base model and one 25-stage stream; given HERD's forgetting of 0.010 and fixed per-stage cost (\S\ref{sec:experiment-results}), we plan to test longer streams with more task families.

\paragraph{Learning from user harnesses.}
Our long-term goal is a continual learner that trains recursively on a user's own harness, the prompts, skills, and memories they already write (\S\ref{sec:introduction}), and so internalizes their organization's procedures, domain knowledge, and codebases.

\paragraph{Beyond verified answers and known boundaries.}
User harnesses supply neither verified answers nor task boundaries.
To reduce the dependence on $y^*$, we are considering: (i) computing the residual and the guard on an accepted target other than $y^*$, such as a user correction, a rubric, or a majority answer from several harness rollouts; (ii) opening a stage when the mean residual of incoming records rises and closing it when it falls; and (iii) routing skills by retrieval over the record rather than by a template prefix, so that one stage may hold several tasks.

\paragraph{Recursive self-improvement.}
Because the harness HERD evolves at one stage is distilled into the weights that evolve it at the next, the loop could extend to recursive self-improvement \citep{zhang-etal-2026-dgm}: (i) letting the current checkpoint rather than the frozen $\theta_0$ write the skill (\S\ref{sec:recursive-bounded-harness}), so that a model that has internalized one stage writes a better skill for the next; (ii) placing the drift budgets, window size, and record selection in the harness, where the model proposes changes adopted only after a controlled commit; and (iii) applying the same residual test and drift audit to every self-proposed change, with rollback when a gate fails.

\end{document}